\documentclass{article}
\usepackage{arxiv}
\usepackage[square,numbers]{natbib}
\usepackage[utf8]{inputenc} 
\usepackage[T1]{fontenc}    
\usepackage{hyperref}       
\usepackage{url}            
\usepackage{booktabs}       
\usepackage{amsfonts}       
\usepackage{nicefrac}       
\usepackage{microtype}      
\usepackage{lipsum}		
\usepackage{graphicx}
\usepackage{doi}
\usepackage{tabularx}
\usepackage{subcaption}
\usepackage{amsmath}
\usepackage{mdframed}
\usepackage{longtable}
\newcolumntype{P}[1]{>{\raggedright\arraybackslash}p{#1}}
\usepackage{comment}
\usepackage{csquotes}
\usepackage[table]{xcolor}

\usepackage{multirow}
\usepackage{array}
\usepackage{tcolorbox} 
\usepackage{listings}
\usepackage{framed}
\newenvironment{highlights}{\begin{framed}}{\end{framed}}
\usepackage{enumitem}
\usepackage{tikz}
\usetikzlibrary{arrows.meta,positioning,fit}
\usepackage{newtxtext}
\usepackage{soul}

\usepackage{fancyvrb}

\usepackage{tikz}
\usetikzlibrary{arrows.meta,positioning,fit}
\usepackage{filecontents}
\lstdefinestyle{pythonstyle}{
    language=Python,
    backgroundcolor=\color{black!5},
    commentstyle=\color{green!40!black},
    keywordstyle=\color{blue},
    stringstyle=\color{purple},
    numberstyle=\scriptsize\color{black!60},
    basicstyle=\ttfamily\fontsize{6}{7}\selectfont,
    breakatwhitespace=false,
    breaklines=true,
    captionpos=b,
    keepspaces=true,
    numbers=left,
    numbersep=4pt,
    showspaces=false,
    showstringspaces=false,
    showtabs=false,
    tabsize=2,
    frame=single,
    framerule=0.4pt
}

\DefineVerbatimEnvironment{HighlightVerbatim}{Verbatim}
{fontsize=\fontsize{8pt}{10pt}\selectfont, commandchars=\\\{\},formatcom=\setlength{\parindent}{0pt}}

\title{Compiling Prompts, Not Crafting Them: A Reproducible Workflow for AI-Assisted Evidence Synthesis}

\author{ 
\href{https://orcid.org/0000-0001-9416-1435}{\includegraphics[scale=0.06]{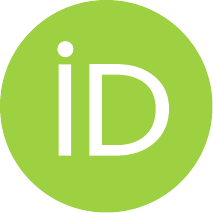}\hspace{1mm}Teo ~Susnjak} \thanks{Corresponding author: t.susnjak@massey.ac.nz} \\
	School of Mathematical and Computational Sciences\\
	Massey University\\
	Albany, New Zealand \\
}
\renewcommand{\shorttitle}{Compiling Prompts, Not Crafting Them}

\begin{document}
\maketitle

\begin{abstract}
Large language models (LLMs) offer significant potential to accelerate systematic literature reviews (SLRs), yet current approaches often rely on brittle, manually crafted prompts that compromise reliability and reproducibility. This fragility undermines scientific confidence in LLM-assisted evidence synthesis. In response, this work adapts recent advances in declarative prompt optimisation, developed for general-purpose LLM applications, and demonstrates their applicability to the domain of SLR automation. This research proposes a structured, domain-specific framework that embeds task declarations, test suites, and automated prompt tuning into a reproducible SLR workflow. These emerging methods are translated into a concrete blueprint with working code examples, enabling researchers to construct verifiable LLM pipelines that align with established principles of transparency and rigour in evidence synthesis. This is a novel application of such approaches to SLR pipelines.

\end{abstract}


\begin{highlights}

\subsection*{What is already known}
\begin{itemize}
    \item Systematic literature reviews (SLRs) are foundational for evidence-based practice but are notoriously slow and resource-intensive.
    \item Large language models (LLMs) show significant promise for automating SLR tasks, but their performance is highly sensitive to the phrasing of input prompts.
    \item This "prompt fragility" makes LLM-assisted workflows unreliable, difficult to reproduce, and raises concerns about their scientific validity.
\end{itemize}

\subsection*{What is new}
\begin{itemize}
    \item This paper introduces a declarative framework that adapts state-of-the-art prompt optimisation techniques from the general AI field and applies them specifically to the domain of SLR automation.
    \item It replaces manual, ad-hoc "prompt alchemy" with a rigorous, four-step programmatic process: (1) formally defining the research goal, (2) codifying the quality standard with data, (3) automatically compiling an optimal prompt, and (4) packaging the result as a verifiable digital artefact.
    \item The study provides both a conceptual blueprint and a functional, code example, demonstrating a practical pathway to building robust and reproducible LLM pipelines for evidence synthesis.
\end{itemize}

\subsection*{Potential impact for research automation and synthesis researchers}
\begin{itemize}
    \item This work provides researchers and methodologists with a clear, actionable methodology to harness the speed of LLMs without sacrificing the scientific rigour and reproducibility that are cornerstones of evidence synthesis.
    \item The proposed framework offers a path toward establishing new standards for transparency and auditability in AI-assisted reviews, allowing others to verify and replicate automated steps with precision.
    \item It lays the groundwork for a future ecosystem of modular, verifiable, and reusable AI components for all stages of an SLR, empowering the community to build more trustworthy and efficient tools for research synthesis.
\end{itemize}

\end{highlights}

\section{Introduction}
Large language models (LLMs) now offer a promising path for automating systematic literature reviews (SLRs) \cite{lieberum2025scoping}. Recent studies show that models have remarkable potential to automate all phases of an SLR process, from abstracts screening, data extraction, quality assessment, through to evidence syntheses with promising accuracy \cite{lieberum2025scoping,Susnjak2025Automating}. Despite progress, serious concerns persist. LLM outputs can change significantly with small variations in LLM prompts \cite{sclar2023quantifying}. Model updates can invalidate previously crafted prompts and break reliable pipelines, while cross-model behaviour and accuracy diverges significantly using identical prompts \cite{li2025levelautomationgoodenough}. These LLM sensitivities and fragilities erode trust and widen the reproducibility gap in SLRs and scientific endeavours more broadly. Therefore, there is a need to explore more systematic approaches to the automation of SLRs that ensure reliability and repeatability when employing LLMs for evidence synthesis.

\section{Prompt Engineering Crisis}
The remarkable reasoning abilities of LLMs have resulted in natural language (NL) becoming framed as a new \textit{programming language}\cite{kearns2023responsible}, implying that NLs have become the primary interface for instructing AI systems, with generative models acting as \textit{compilers} that translate these instructions into correct actions or executable code.
However, this ignores the consequences of the ambiguity of NLs which do not have the  rigidity and precision of highly syntactic programming languages. Humans rely on context to resolve meaning while programming removes ambiguity and assumptions through precise instructions. Treating LLM prompts as code collapses this difference. Since LLMs do not have the ability to produce same outputs for identical inputs, this results in the current state where \textit{prompt engineering} has transformed into \textit{prompt alchemy}. LLM outputs are highly non-deterministic, influenced by \textit{ad hoc} prompt design choices, sequencing of instructions and slight rephrasings, as well as luck \cite{razavi2025benchmarking}. Subtle variations in prompt formats can result in differences as large as 76 accuracy points \cite{sclar2023quantifying} on tasks from the Super-NaturalInstructions benchmark. The evidence of this brittleness is also growing in the SLR automation field \cite{staudinger2024reproducibility,lieberum2025scoping}. While studies identify LLM fragilities, none propose a reproducible, programmatic remedy offering deterministic and more accurate outputs.  
Table \ref{tab:variability} summarises recent works using LLMs to automate different SLR phases, revealing their variability.

\begin{table}[hbtp]
\footnotesize
\centering
\begin{tabular}{r p{0.55\linewidth}}
\toprule
\textbf{SLR Phase \& Study} & \textbf{Prompt Engineering Fragility Example} \\
\midrule

\multicolumn{2}{l}{\textbf{Screening}} \\

\citet{shah2024efficacy} & Accuracy fluctuated by up to 28.3\% across prompts \\

\citet{dennstadt2024title} & Sensitivity/specificity swung from 94.5\%/31.8\% (Flan‑T5) to 81.9\%/75.2\% (Mixtral) across LLM families and prompts. \\

\citet{cao2024prompting} & Reordering or omitting few‑shot exemplars altered include/exclude decisions for ~15\% of abstracts. \\

\citet{trad2025streamlining} & Tightening exclusion thresholds in prompts dropped abstract‑error‑rate (AER) from 72.1\% to 50.7\%. \\

\citet{cao2025development}   & Zero‑shot prompting achieved sensitivity 16.7–87.5\%, vs. optimised prompt 86.7–100.0\%. \\

\midrule

\multicolumn{2}{l}{\textbf{Data Extraction}} \\
\citet{cao2025otto} & Accuracy swung by ~15 percentage points across different prompts. \\
\citet{lai2025language} & LLM‑assisted extraction rose from 95.1\% to 97.9\%,  driven by optimised prompts\\
\citet{li2025levelautomationgoodenough} & Recall for extracting study details varied between 64\% and 92\% across different LLMs and prompting strategies. \\
\citet{khraisha2024can} & Accuracy spanned 60\% to near‑perfect levels (>95\%) depending on prompts \\
\midrule

\multicolumn{2}{l}{\textbf{Risk of Bias / Quality Assessment}} \\

\citet{wang2024prompt} & Agreement with clinical guidelines acutely ranged from kappa = –0.002 to 0.984 under different prompting styles, compromising compliance judgments. \\

\citet{Honghao2024assessing} 
  & RoB 1 accuracy shifted from 84.5\% to 89.5\% across different LLMs and prompt configurations; domain accuracies swung 56.7–98.3\%. \\

\citet{lai2025language} 
  & LLM‑only RoB 1 accuracy was 95.7–96.9\%; domain scores varied from 87.9\% to 100\%. \\

\citet{Eisele2025exploring} 
  & Overall RoB2 LLM agreement was 41\% (kappa=0.22; domain kappa range 0.10–0.31), varying with prompt phrasing and domain focus. \\

\bottomrule
\end{tabular}
\caption{Examples of LLM prompt‑induced performance swings in SLR tasks}
\label{tab:variability}
\end{table}

\section{A Declarative Framework for Reliable and Reproducible SLR Automation}

The brittleness of prompt engineering requires a paradigm shift from \textit{ad hoc} design approaches to programmatic rigour. 
This work proposes the use of \textit{declarative prompt tuning} approaches for future SLR automation research, inspired by recent advances in prompt optimisation such as the DSPy\cite{opsahl2024optimizing}, GRPO\cite{shao2024deepseekmath} and GEPA\cite{agrawal2025gepa} frameworks. 
This study presents a return to a programmatic, and a declarative paradigm specifically, aiming to restore reproducibility in LLM-driven SLR research by decoupling the researcher's scientific intent (the ``what''), from the model's specific implementation (the ``how''). Instead of relying on fragile, hand-crafted prompts, this approach treats LLM workflows for SLR tasks as language model (LM) \textit{programs} that can be compiled. This compilation entails an automated process that systematically searches for a high-performing LLM-agnostic prompt configuration that satisfies a predefined quality standard or accuracy requirement. This methodology is operationalised through four key components depicted and described in Figure \ref{fig:framework} and applicable to all stages of an SLR process. For a concrete illustration, the framework is translated to an example for the abstract screening process below:

\begin{figure}[hbt]
        \centering
        \includegraphics[width=1\linewidth]{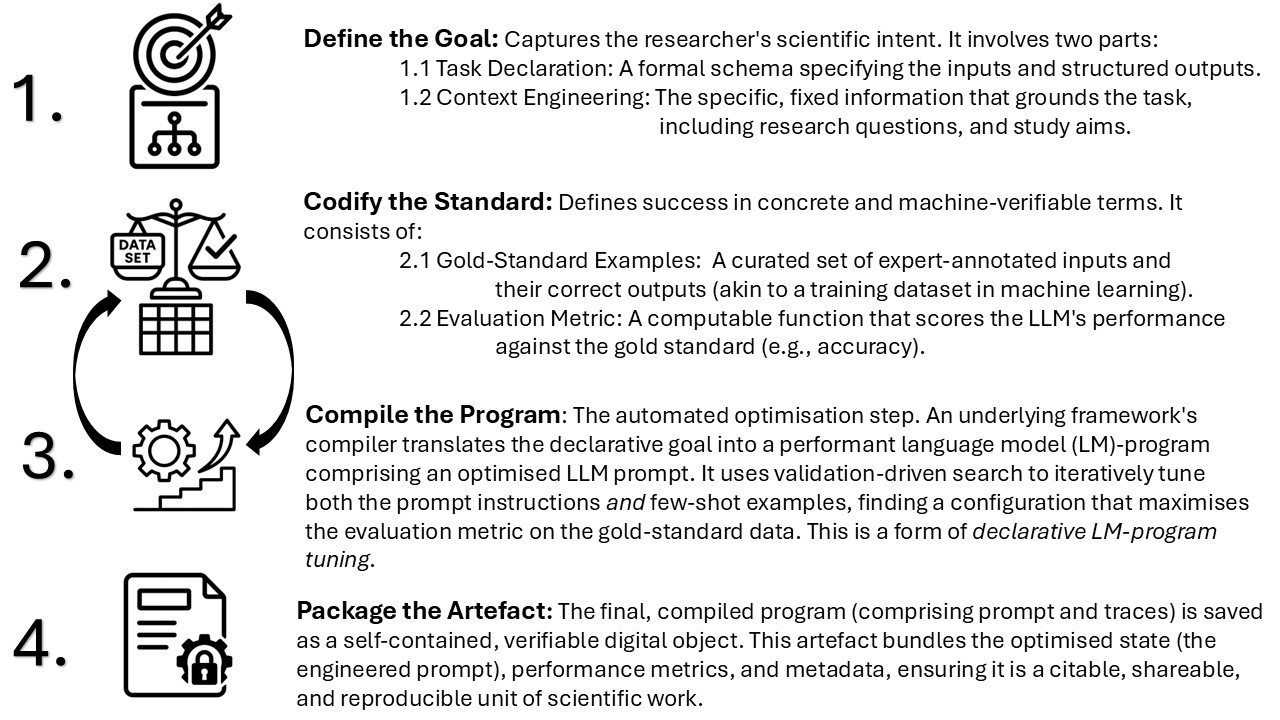}
        \caption{Four Components of the Declarative Language Model-program Tuning Framework}
        \label{fig:framework}
    \end{figure}
    
\begin{tcolorbox}[
    colback=blue!5!white,
    colframe=blue!75!black,
    fonttitle=\bfseries,
    fontupper=\small,
    title=Box 1: Blueprint for a Verifiable Screening Module via Declarative LM-program Tuning,
    label={box:blueprint}
]

\vspace{1em}

\textbf{Abstract Screening Example}

\begin{description}
    \item[\textbf{1. Define the Goal:}] Screening abstracts for inclusion.
  \begin{itemize}
    \item \textbf{Task Declaration} Inputs follow a fixed schema \{\texttt{title}, \texttt{abstract}, \texttt{keywords}\}. The label space is \{\texttt{Include}, \texttt{Exclude}, \texttt{Unsure}\}. \texttt{Unsure} is treated as \texttt{Include} for safety or routed to human review, and this policy is fixed in the spec.
    \item \textbf{Context Engineering} A versioned context file states the PICO criteria, study designs in scope, and the review questions.
  \end{itemize}

  \item[\textbf{2. Codify the Standard}] Define a machine-testable target.
  \begin{itemize}
    \item \textbf{Gold-Standard Examples} Curate $N$ expert-labeled abstracts not part of the study, representing each possible classification outcome.
    \item \textbf{Evaluation Metric} Primary metric is accuracy.
  \end{itemize}

  \item[\textbf{3. Compile the Program}] Run controlled search over prompts and exemplars.
  \begin{itemize}
    \item The compiler explores instruction templates and up to $k$ few-shot exemplars under a pinned model build with \texttt{temperature=0}, fixed \texttt{seed}, and a set budget $B$ evaluations. All runs log hashes of data, prompts, model ID, and decoding parameters.
  \end{itemize}

  \item[\textbf{4. Package the Artefact}] Emit a shareable, auditable bundle.
  \begin{itemize}
    \item The bundle contains \texttt{config.yaml} (task spec and run controls), \texttt{prompt.txt} and \texttt{exemplars.json}, \texttt{metrics.json} with the test-set results, and a run log. A short mapping lists which PRISMA items the bundle supports, such as protocol transparency and decision traceability. Results are verifiable and recomputable under a pinned environment.
  \end{itemize}
\end{description}

\end{tcolorbox}

The conceptual blueprint detailed in Box 1 can be directly implemented using modern programmatic LLM frameworks. To further demonstrate how this can be operationalised, a functional Python implementation of the above illustrated abstract screening module using the DSPy (MIPROv2) library is provided in Appendix A. \href{https://colab.research.google.com/drive/1y24xxCDW06TKTnlKmhe30zuYBQa-szdY?usp=sharing}{A MIPROv2 example} and an equivalent example using the latest \href{https://colab.research.google.com/drive/1bQJbeC6bNDQw_OZbSgv3RoYoFnUkLcYy?usp=sharing}{GEPA implementation} are made available on Google Colab. The code example in the Appendix A demonstrates a declarative  goal definition, a gold-standard dataset and a metric function to codify the standard, an optimiser that compiles the LM-program, as well as how the verifiable artefact is packaged and used to make classifications on new abstracts. 
This automated compilation process is analogous to the ``hyperparameter tuning'' in machine learning, in that it uses a validation dataset and a metric to systematically search for an optimal configuration. However, instead of tuning architectural parameters, this framework tunes NL artefacts which comprise the instructions and few-shot examples that guide a fixed, pre-trained model. This positions the method as a rigorous, data-driven alternative to manual prompt engineering, meeting the high scientific standards required for evidence synthesis.
This process is can be seen as analogous to the ``hyperparameter tuning'' process in machine learning that ensures that ``near''-optimal parameters are derived for a given model and dataset prior to use; thus, this approach is rigorous and meets high scientific requirements that manual prompt engineering does not. 
This probing study calls on researchers in the SLR automation field to empirically investigate this approach and available supporting framework implementations, and thus provide a tangible pathway for researchers to adopt this rigorous and reproducible methodology for fully automating SLRs in a PRISMA-compliant manner.

\section{Conclusion}
LLMs offer a compelling opportunity to streamline the labour-intensive processes involved in systematic literature reviews, but the current reliance on brittle prompt engineering introduces unacceptable risks to reproducibility and scientific rigour. This exploratory study proposes an alternative: a declarative framework for SLR automation that adapts recent advances in prompt optimisation into a structured, testable, and version-controlled workflow. 
The contribution here lies in applying declarative prompt tuning approaches, originally developed for general LLM tasks—to the domain of SLR automation, and demonstrating their utility through a proof-of-concept, reproducible implementation. To the best of current knowledge, this represents the first application of such declarative techniques to evidence synthesis workflows.
The contribution here lies in adapting emerging declarative prompt optimisation techniques to SLR automation and demonstrating their utility through a reproducible implementation. 
This prototyping work provides both a conceptual blueprint and a working implementation, illustrating how LLM-assisted evidence synthesis can evolve from fragile, \textit{ad hoc} prompting into verifiable, auditable, and modular pipelines. Future work should fully test and expand this approach to other SLR stages, integrate it into standard reporting frameworks, and foster a community-driven ecosystem of reusable components for transparent, AI-enabled reviews.

\bibliographystyle{unsrtnat}


\appendix

\section*{Appendix A: Technical Example of a Screening Module for Prompt Optimisation}
\label{app:code}

The conceptual blueprint described in Box 1 can be implemented using programmatic LLM tools like DSPy. The following Python code provides a minimal functional example of how an \texttt{AbstractScreening} module would be structured, tested, and compiled into a verifiable, reproducible digital artefact, translating the declarative framework to the corresponding components of the underlying tool. A more complete example of the code below can be found \href{https://colab.research.google.com/drive/1y24xxCDW06TKTnlKmhe30zuYBQa-szdY?usp=sharing}{ on a Google Colab notebook}.

\begin{lstlisting}[
    style=pythonstyle,
    caption={Python implementation of the proposed framework for an abstract screening module, structured into the four core components: Goal Definition, Standard Codification, Program Compilation, and Artefact Packaging},
    label={lst:dspy_example}
]
import dspy
from typing import Literal

# Define the parameters you want to pass to the chosen LLM
# Fix parameters to maximise determnistic responses
model_params = {
    "temperature": 0.0,
    "top_p": 1.0,      
    "seed": 42,         
    "max_tokens": 1024  
}

# LLM and DSPy configuration
# Pass the parameters by unpacking the dictionary using **
lm = dspy.LM(
    "openai/gpt-4o-mini", 
    api_key="<YOUR_API_KEY>", 
    **model_params
)

# LLM and DSPy configuration (e.g., dspy.configure) is assumed.
dspy.settings.configure(lm=lm)

## --- COMPONENT 1: Define the Goal --- ##
# This includes both the Task Declaration (the Signature) and the Context Engineering (the specific criteria for this review).

# 1a: Task Declaration
class ScreenAbstract(dspy.Signature):
    """Screens an abstract based on predefined research context and criteria."""
    
    # Input fields for context and the item to be screened.
    criteria: str = dspy.InputField(desc="Fixed PICOS criteria for the review")
    study_aims: str = dspy.InputField(desc="The primary objective of the study")
    research_question: str = dspy.InputField(desc="The research question guiding the review")
    abstract: str = dspy.InputField(desc="The abstract text to be screened")
    
    # Output fields with structured constraints.
    decision: Literal["Include", "Exclude", "Unsure"] = dspy.OutputField()
    reasoning: str = dspy.OutputField(desc="Brief justification for the decision, referencing the criteria")
    confidence: float = dspy.OutputField(desc="A score from 0.0 to 1.0 indicating the model's confidence")

# 1b: Context Engineering
CRITERIA = (
    "Population: adults 18-65 with major depressive disorder. "
    "Intervention: digital CBT via app or web. "
    "Design: RCTs or non-inferiority trials. "
    "Outcome: validated depressive symptom scale."
)
STUDY_AIMS = "To determine if digital CBT reduces depressive symptoms in adults with MDD."
RESEARCH_QUESTION = (
    "In adults aged 18-65 with MDD, does digital CBT, compared with usual care, "
    "reduce depressive symptoms measured using validated scales?"
)

## --- COMPONENT 2: Codify the Standard --- ##
# This establishes the quality bar with gold-standard examples and a success metric.

# 2a: Gold-Standard Examples
gold_standard = [
    dspy.Example(
        criteria=CRITERIA, study_aims=STUDY_AIMS, research_question=RESEARCH_QUESTION,
        abstract="An RCT randomized 120 adults 25-60 with DSM-5 MDD to digital CBT vs usual care;
                    PHQ-9 reported at 12 weeks...",
        decision="Include",
        reasoning="The abstract describes an RCT in the correct population using digital CBT and a valid 
                    outcome measure."
    ).with_inputs("criteria", "study_aims", "research_question", "abstract"),
    dspy.Example(
        criteria=CRITERIA, study_aims=STUDY_AIMS, research_question=RESEARCH_QUESTION,
        abstract="A cohort study of a web-based CBT tool among teenagers aged 13-17; no control group...",
        decision="Exclude",
        reasoning="This is not an RCT and the population (teenagers) is incorrect."
    ).with_inputs("criteria", "study_aims", "research_question", "abstract"),
    dspy.Example(
        criteria=CRITERIA, study_aims=STUDY_AIMS, research_question=RESEARCH_QUESTION,
        abstract="Participants described as adults with depressive symptoms were assigned to a web program 
        combining CBT content with mindfulness and peer support versus usual care with PHQ-9 at 10 weeks.
        Formal MDD diagnosis and randomization are not stated and the upper age bound is not clear.",
        decision="Unsure",
        reasoning="Intervention fidelity to CBT and confirmation of MDD are uncertain and allocation and age
        bounds are unclear which prevents a confident decision."
    ).with_inputs("criteria","study_aims","research_question","abstract"),
    
    # A real-world set would contain 10-50 high-quality examples.
    # ...
]

# 2b: Evaluation Metric
def screening_accuracy_metric(gold: dspy.Example, pred: dspy.Prediction, trace=None) -> bool:
    """A simple metric: the predicted decision must match the gold-standard decision."""
    return gold.decision == pred.decision

## --- COMPONENT 3: Compile the Program --- ##
# An optimizer tunes a program (e.g., a ChainOfThought module) to maximize the chosen metric.
# This is declarative LM-program tuning.


optimizer = dspy.MIPROv2(metric=screening_accuracy_metric, num_threads=4)
compiled_screener = optimizer.compile(
    valset=gold_standard[:10],      # validation set with a simple approach assuming one has 10 samples
    trainset=gold_standard[10:],    # training set assuming one has more than 10 samples
    student=dspy.ChainOfThought(ScreenAbstract),
    max_bootstrapped_demos=0,
    max_labeled_demos=1,
    minibatch=False,
    requires_permission_to_run=False
)

## --- COMPONENT 4: Package the Artefact --- ##
# The compiled program's state is saved to a file, creating a reproducible artefact.

compiled_screener.save(path="screen_abstract_v1.json")

# --- Usage Example of the Artefact ---
# A collaborator can load the artefact and achieve identical performance.
loaded_screener = dspy.ChainOfThought(ScreenAbstract)
loaded_screener.load("screen_abstract_v1.json")

new_abstract = (
"This randomized controlled trial enrolled 180 adults aged 28-62 diagnosed with major depressive disorder.
 Participants received app-based CBT vs wait-list control. Depressive symptoms were measured using PHQ-9 at 8 weeks."
)

# Run prediction with the loaded, optimized program.
prediction = loaded_screener(
    criteria=CRITERIA,
    study_aims=STUDY_AIMS,
    research_question=RESEARCH_QUESTION,
    abstract=new_abstract
)
print("Decision:", prediction.decision)
print("Reasoning:", prediction.reasoning)
print("Confidence:", prediction.confidence)

# --- Output Example ---
Decision: Include
Reasoning: The study meets all the PICOS criteria: it involves adults with major depressive disorder (Population), uses digital CBT via an app (Intervention), is a randomized controlled trial (Design), and measures outcomes using a validated depressive symptom scale (Outcome). The study also aligns with the study aims as it investigates the effect of digital CBT on depressive symptoms in adults with MDD.
Confidence: 1.0
\end{lstlisting}

\end{document}